\documentclass[preprint,12pt]{elsarticle}
\usepackage[margin=1in]{geometry}
\usepackage[table]{xcolor}
\usepackage{amsmath,amssymb,amsfonts}
\usepackage{algorithm}
\usepackage{algorithmic}
\usepackage{graphicx}
\usepackage{enumitem}

\newcommand{\safeincludegraphics}[2][]{%
  \IfFileExists{#2}{\includegraphics[#1]{#2}}{%
    \fbox{\begin{minipage}[c][0.18\textheight][c]{0.9\linewidth}\centering Missing figure file:\\ \texttt{\detokenize{#2}}\end{minipage}}%
  }%
}
\usepackage{textcomp}
\usepackage{tabularx}
\usepackage{booktabs}
\usepackage{multirow}
\usepackage{array}
\usepackage{url}
\usepackage{hyperref}

\makeatletter
\def\ps@pprintTitle{%
  \let\@oddhead\@empty
  \let\@evenhead\@empty
  \let\@oddfoot\@empty
  \let\@evenfoot\@empty
}
\makeatother

\begin{document}

\begin{frontmatter}

\title{DualMiT-Net: Local-Global Transformer-Convolutional Fusion for Breast Mass Segmentation in Mammographic Regions of Interest}

\author[inst1]{Alibek Kamiluly}
\author[inst1]{Milana Muratova}
\author[inst2]{Yash Patel}
\author[inst1]{Fan Li\corref{cor1}}
\cortext[cor1]{Corresponding author.}

\address[inst1]{Department of Mathematics and Computer Science,
Lawrence Technological University}

\address[inst2]{College of Business and Information Technology,
Lawrence Technological University}

\begin{abstract}
Breast mass segmentation is an important step in computer-aided mammography, but it remains difficult because masses can have low contrast, irregular shapes, and boundaries that blend with surrounding breast tissue. To address this problem, we present DualMiT-Net, a dual-branch network that uses both a focused view of the mass and a wider view of the surrounding tissue. The local branch uses a Mix Transformer (MiT-B5) encoder to learn mass shape, texture, and boundary information, while the global branch uses an EfficientNet-B5 encoder to learn surrounding breast context. Features from the two branches are shared at the deeper encoder levels and are then progressively fused in a single decoder. A spatial gate controls how much global information is added during decoding. We also evaluated four input representations and selected a percentile-windowed mammogram combined with a Gabor texture response. The model was trained and evaluated on the mass subset of the Curated Breast Imaging Subset of the Digital Database for Screening Mammography (CBIS-DDSM) using a patient-level split. Across three training runs, DualMiT-Net with exponential moving average weights achieved a mean Dice coefficient of $0.9375 \pm 0.0011$ and a mean Intersection over Union  of $0.8834 \pm 0.0020$. It also achieved better Dice and IoU scores than six standard encoder-decoder baselines trained using the same data and training settings. These results show that combining local mass information with wider breast context can provide accurate and consistent breast mass segmentation.
\end{abstract}


\end{frontmatter}

\section{Introduction}
\label{sec:introduction}

Breast cancer remains a major global health burden. The American Cancer Society projects an estimated 2,041,910 new cancer cases in the United States for 2025, of which 316,950 are breast cancers in women, approximately 32\% of all new female cancer diagnoses~\cite{siegel2025cancer}. Early detection remains important because the five-year relative survival rate exceeds 99\% for localized breast cancer~\cite{acs2024facts}. Semantic segmentation, which assigns a class label to every pixel of an image, provides a means of delineating breast masses precisely and thereby supports the characterization of a lesion once it has been identified.

Accurate breast mass segmentation remains difficult. Mammograms can have low contrast between a mass and the surrounding tissue, image noise, and overlapping breast structures that make the boundary hard to see~\cite{sun2020aunet,wang2020ammsp,xu2022arfnet}. Manual mass contours can also vary between readers~\cite{sahiner2001characterization}. Reading mammograms is time-sensitive: a large prospective screening study reported an average reading time of about one minute per case for two-dimensional full-field digital mammography and about two minutes for digital breast tomosynthesis~\cite{partridge2024reading}. Automated segmentation may therefore help provide a consistent lesion outline for review.

Deep learning has greatly improved medical image segmentation. Fully convolutional networks~\cite{long2015fully} introduced end-to-end pixel prediction, but their output could lose fine spatial detail. U-Net~\cite{ronneberger2015u} addressed this problem by using skip connections to pass detailed features from the encoder to the decoder. Later models such as U-Net++~\cite{zhou2018unet++} and attention-gated U-Net~\cite{schlemper2019attention} changed how these features are combined. DeepLabv3+~\cite{chen2018encoder} used atrous spatial pyramid pooling to capture information at several scales, while ResU-Net~\cite{diakogiannis2020resunet} added residual connections to support deeper networks.

Convolutional networks are effective at learning local image patterns, but it can be harder for them to model relationships between distant regions of an image~\cite{wang2018non}. Vision Transformers (ViTs)~\cite{dosovitskiy2020image} use self-attention to connect information across the image. Hybrid transformer-convolutional models have also been successful for segmentation. TransUNet~\cite{chen2021transunet} combines a ViT with a U-Net structure, while SegFormer~\cite{xie2021segformer} uses the Mix Transformer (MiT) encoder to produce multi-scale features efficiently. Most of these models still use only one image view. For breast mass segmentation, one view must then provide both fine boundary detail and enough surrounding tissue for context. A tight crop is useful for the boundary, while a wider view is useful for context.

This paper presents DualMiT-Net, a dual-branch encoder-decoder model for mass segmentation in lesion-centered mammographic regions of interest (ROIs). The local branch receives a lesion-centered ROI and uses a MiT-B5 encoder to learn mass shape, texture, and boundary information. The global branch receives a wider view of the same area and uses an EfficientNet-B5 encoder to learn information from the surrounding breast tissue. The two branches exchange information at their deepest encoder levels, and their features are then progressively fused in a single decoder to generate the final segmentation mask. The main contributions of this work are as follows.

\begin{enumerate}[leftmargin=*, labelsep=0.5em]

\item \emph{Dual-view local-global architecture:} DualMiT-Net jointly uses a focused lesion ROI and a wider contextual view to capture both mass boundary detail and surrounding breast-tissue information.

\item \emph{Deep cross-branch and spatially gated fusion:} The model exchanges local and global features through deep feature sharing and selectively incorporates contextual information during decoding.

\item \emph{Texture-aware input and matched evaluation:} A percentile-windowed Gabor representation is evaluated together with DualMiT-Net, which achieves a Dice of $0.9375$ and an IoU of $0.8834$, outperforming six standard segmentation baselines under the same experimental protocol.
\end{enumerate}

The remainder of this paper is organized as follows. Section~\ref{sec:related_work} reviews related work in mammographic image segmentation. Section~\ref{sec:materials_methods} describes the dataset, preprocessing pipeline, network architecture, loss function, evaluation metrics, and experimental protocol. Section~\ref{sec:results} reports the experimental results, Section~\ref{sec:discussion} interprets them and states the limitations of the study, and Section~\ref{sec:conclusion} concludes.

\section{Related Work}
\label{sec:related_work}

\subsection{Convolutional Neural Network (CNN)-Based Breast Mass Segmentation}
\label{subsec:cnn_segmentation}

Convolutional Neural Networks (CNNs) have been widely used for breast mass segmentation because they can learn image features directly from mammograms. U-Net~\cite{ronneberger2015u} became a common foundation for medical image segmentation because its encoder-decoder structure combines high-level features with spatial information from earlier layers. Many breast mass segmentation methods have extended this structure to better handle low contrast, irregular boundaries, and the large variation in mass size and shape.

Several studies have evaluated breast mass segmentation on the Curated Breast Imaging Subset of the Digital Database for Screening Mammography (CBIS-DDSM). AUNet~\cite{sun2020aunet} introduced attention-guided dense upsampling for whole-mammogram segmentation. AM-MSP-cGAN~\cite{wang2020ammsp} used an adversarial framework with attention and multi-scale pooling. MTLNet~\cite{hou2021mtlnet} combined attention with multi-task learning so that segmentation, localization, and classification could share learned features. ARF-Net~\cite{xu2022arfnet} introduced adaptive receptive-field learning to better handle masses with different sizes. Connected-ResUNets~\cite{baccouche2021connected} connected residual U-Net stages to improve feature reuse and segmentation refinement. These studies show that attention, multi-scale features, receptive-field design, and adversarial learning can all improve breast mass segmentation.

\subsection{Transformer-Based and Local-Global Segmentation}
\label{subsec:transformer_segmentation}

Vision Transformers (ViTs)~\cite{dosovitskiy2020image} use self-attention to model relationships between distant image regions. Hybrid architectures such as TransUNet~\cite{chen2021transunet} combine transformer-based global modeling with the spatial detail provided by convolutional or U-Net-like components. SegFormer~\cite{xie2021segformer} uses the Mix Transformer (MiT) encoder to produce multi-scale features with efficient self-attention. This is relevant to breast mass segmentation because the lesion boundary requires fine spatial detail, while the surrounding tissue can provide additional context.

Transformer and multi-view ideas have also been explored directly in mammography. YOLO-LOGO~\cite{su2022yolo} combines mass detection with local and global transformer branches for segmentation in digital mammograms. Ma and Peng~\cite{ma2024crossview} used complementary cranio-caudal and mediolateral oblique mammographic views in a cross-view variational autoencoder and reported strong segmentation performance on CBIS-DDSM. These studies support the use of broader contextual information, but they use different ways of defining and combining local, global, or cross-view information.

Previous breast mass segmentation methods have addressed different parts of the segmentation problem. Attention-based and multi-scale methods such as AUNet, AM-MSP-cGAN, MTLNet, and ARF-Net focus on feature selection or on handling masses with different sizes and shapes. Connected-ResUNets uses connected residual encoder-decoder stages to refine the segmentation result. Transformer-based approaches provide a wider receptive field, while YOLO-LOGO combines local and global transformer features. Ma and Peng use complementary mammographic views to include additional information during segmentation.

DualMiT-Net differs from these approaches by using two aligned fields of view from the same lesion. The local view provides detailed information about mass shape, texture, and boundary structure, while the wider global view includes more of the surrounding breast tissue. The two views are processed by separate encoders, exchange information at deeper feature levels, and are progressively fused in a single decoder. A spatial gate controls how much information from the wider view is added during decoding. This design allows the model to preserve local boundary information while also using surrounding tissue as additional context.

Table~\ref{tab:literature_summary} summarizes breast mass segmentation methods evaluated on CBIS-DDSM. The table includes the main design idea of each method together with the reported Dice and Intersection over Union (IoU) values when available. A dash indicates that the original study did not report IoU for the CBIS-DDSM experiment. Jaccard values are shown in the IoU column because the Jaccard index and IoU represent the same overlap measure.

\begin{table*}[!htbp]
\centering
\caption{Summary of selected breast mass segmentation methods evaluated on CBIS-DDSM.}
\label{tab:literature_summary}
\small
\begin{tabularx}{\textwidth}{p{3.6cm} p{3.2cm} p{5cm}X cc}
\toprule
Study & Method & Main idea & Dice & IoU \\
\midrule
AUNet~\cite{sun2020aunet} & Attention-based encoder-decoder & Attention-guided dense upsampling for whole mammograms & 0.8180 & - \\
AM-MSP-cGAN~\cite{wang2020ammsp} & Conditional GAN & Attention and multi-scale pooling with adversarial learning & 0.8449 & - \\
ARF-Net~\cite{xu2022arfnet} & Adaptive receptive-field network & Selects receptive fields for masses with different sizes & 0.8575 & - \\
MTLNet~\cite{hou2021mtlnet} & Attentive multi-task network & Joint segmentation, localization, and classification & 0.8630 & - \\
Connected-ResUNets~\cite{baccouche2021connected} & Connected residual U-Net & Connected U-Net stages with residual learning & 0.8952 & 0.8002 \\
YOLO-LOGO~\cite{su2022yolo} & Local-global transformer & Detection followed by local and global transformer segmentation & 0.7450 & 0.6400 \\
Cross-view VAE~\cite{ma2024crossview} & Cross-view variational autoencoder & Uses complementary mammographic views for segmentation and classification & 0.9246 & 0.8720 \\
\bottomrule
\end{tabularx}
\end{table*}

\section{Materials and Methods}
\label{sec:materials_methods}

\begin{figure*}[!htbp]
  \centering
  \safeincludegraphics[width=\textwidth]{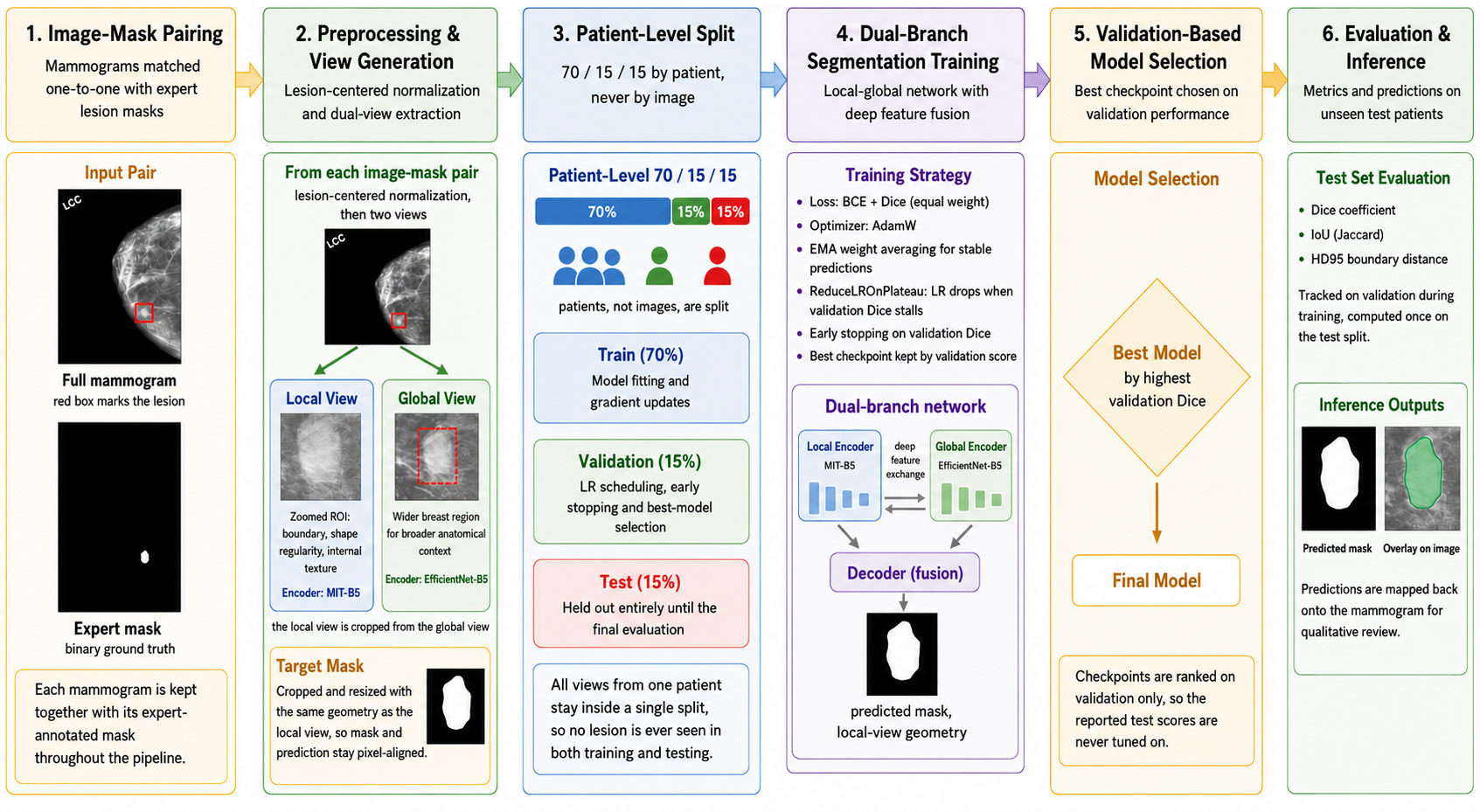}
  \caption{Overview of the proposed segmentation pipeline.}
  \label{fig:overall_pipeline}
\end{figure*}

The proposed framework has two main stages: preprocessing and dual-branch segmentation, as shown in Fig.~\ref{fig:overall_pipeline}. Each mammogram is prepared as two aligned inputs. The local input is centered on the lesion, while the global input contains a wider area of surrounding breast tissue. Both inputs are then passed to the segmentation network.

\subsection{Dataset}
\label{subsec:dataset}

\subsubsection{Data Source}
\label{subsubsec:data_source}

Experiments were conducted on the Curated Breast Imaging Subset of the Digital Database for Screening Mammography (CBIS-DDSM)~\cite{lee2017curated,clark2013cancer}, a publicly available collection of digitized mammographic cases with expert lesion annotations. Each case provides a mammogram together with a corresponding binary lesion mask. Only mass cases exhibiting a valid image-mask correspondence were retained; calcification cases and cases whose mask could not be matched to an image were excluded.

\subsubsection{Patient-Level Partitioning}
\label{subsubsec:split}

Mammography datasets may contain more than one view from the same patient. If images were split independently, related views could appear in both the training and test sets and make performance look better than it really is. To avoid this problem, the dataset was split at the patient level. All images from one patient were kept in only one subset.

Table~\ref{tab:dataset_split} shows the final patient-level split. The training set contains 753 image-mask pairs from 480 patients, the validation set contains 91 pairs from 62 patients, and the test set contains 92 pairs from 64 patients. Image and mask counts match in every subset, confirming that each retained image has one corresponding mask. Some patients contribute more than one mammographic view, which is why the number of image-mask pairs is larger than the number of patients. The validation set was used for model selection and learning-rate scheduling. The test set was used only for the final results reported in Section~\ref{sec:results}. After preprocessing, each image-mask pair produces one local ROI, one wider global image, and one binary target mask.

\begin{table}[!t]%[!htbp]
\centering
\caption{Patient-level distribution of the CBIS-DDSM mass cases used in this study.}
\label{tab:dataset_split}
\begin{tabular}{lrrrr}
\toprule
Subset & Patients & Paired cases & Image files & Mask files \\
\midrule
Training & 480 & 753 & 753 & 753 \\
Validation & 62 & 91 & 91 & 91 \\
Test & 64 & 92 & 92 & 92 \\
\bottomrule
\end{tabular}
\end{table}

\begin{figure}[!htbp]
  \centering
  \safeincludegraphics[width=0.75\textwidth]{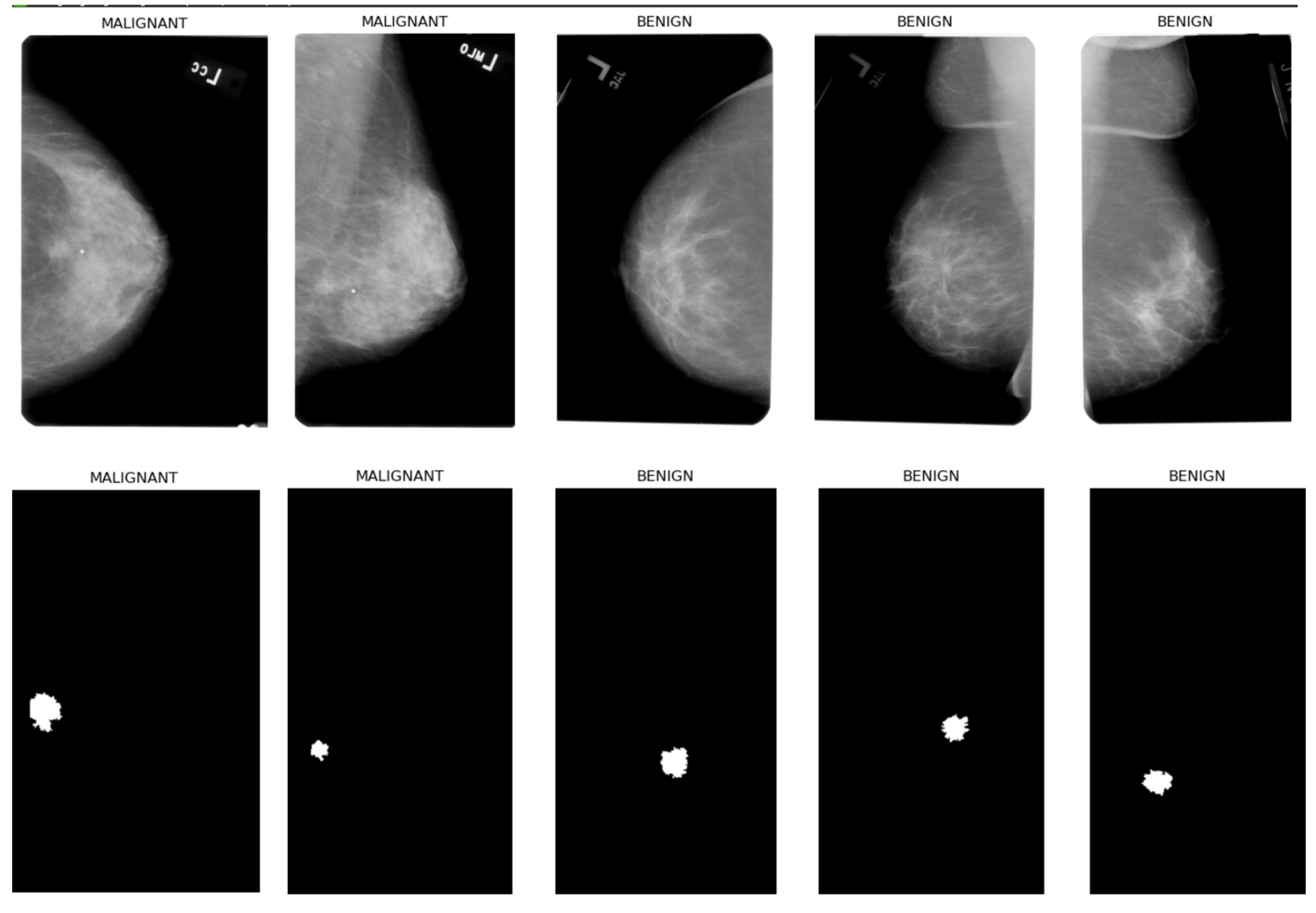}
  \caption{Examples from the CBIS-DDSM mass subset. The top row shows mammograms and the bottom row shows the corresponding expert mass masks.}
  \label{fig:dataset_examples}
\end{figure}

Representative CBIS-DDSM mass cases and their corresponding expert masks are shown in Fig.~\ref{fig:dataset_examples}. The examples illustrate variation in breast appearance, lesion size, lesion location, and mass shape across the dataset.

\subsection{Preprocessing}
\label{subsec:preprocessing}

For each lesion, preprocessing creates two aligned views and one target mask. The lesion position is obtained from the expert annotation mask and is used to define a lesion-centered crop. A tighter local region is extracted around the mass and resized to $256 \times 256$ pixels to preserve lesion shape and boundary detail. A wider region centered on the same lesion is extracted in parallel and resized to $512 \times 512$ pixels so that more of the surrounding breast tissue is retained. The corresponding lesion mask is transformed using the same geometric operations as the local image and is resized to $256 \times 256$ pixels to form the segmentation target. Before the views are extracted, the mammogram and mask are jointly normalized for lesion position, orientation, and scale so that their spatial alignment is preserved. Algorithm~\ref{alg:preprocessing} summarizes the complete preprocessing procedure.

\subsubsection{Intensity standardization and mask preparation}
\label{subsubsec:intensity}

Mammograms were converted to grayscale and normalized by percentile-based intensity windowing. Let $I$ denote the original mammogram. Intensities below a lower percentile and above an upper percentile of the image histogram were clipped, and the clipped image was linearly rescaled to the interval $[0,1]$ to yield the windowed image $I_w$. Percentile-based clipping was used instead of fixed gray-level clipping because image intensity can vary across digitized screen-film mammograms. Lesion masks were binarized such that lesion pixels take the value one and background pixels the value zero. Nearest-neighbor interpolation was applied to all mask resizing operations, so that no intermediate values are introduced at mask boundaries, and bilinear interpolation was applied to image resizing.

\subsubsection{Input representation}
\label{subsubsec:representation}

The encoders used in this work were pretrained on natural images and expect a three-channel input, while a mammogram has only one intensity channel. Instead of copying the same channel three times, one channel is used to add texture information. In the notation below, square brackets mean that the channels are stacked together. The selected input is
\begin{equation}
\mathbf{X} = [\,I_w,\; I_w,\; I_g\,],
\label{eq:representation}
\end{equation}
where $\mathbf{X}$ denotes the three-channel network input; $I_w$ denotes the percentile-windowed mammogram defined in Section~\ref{subsubsec:intensity}; and $I_g$ denotes the Gabor texture response computed from $I_w$ and normalized to $[0,1]$. A Gabor filter responds to oriented texture patterns. This is useful for mammograms because spiculation and margin texture can help describe the mass boundary. The windowed image is repeated in the second channel to preserve compatibility with the pretrained encoder weights. Three alternative representations, evaluated in Section~\ref{subsec:input_ablation}, were also constructed: the windowed image alone, the windowed image combined with Contrast Limited Adaptive Histogram Equalization (CLAHE), and the combination of windowing, CLAHE, and Gabor filtering.

\subsubsection{Local and global view extraction}
\label{subsubsec:views}

For each case, the lesion position was obtained from the expert annotation mask. A lesion-centered local ROI was cropped using the lesion bounding box with a small additional margin and resized to $256 \times 256$ pixels. The corresponding mask region was transformed in the same way and resized to $256 \times 256$ pixels to form the segmentation target. A wider region centered on the same lesion was extracted in parallel and resized to $512 \times 512$ pixels so that more of the surrounding breast tissue was retained. This wider view is processed by the global encoder, and its features are resampled to match the spatial resolution of the decoder features during fusion.

Because the lesion position and extent are obtained from the expert annotation, the network performs segmentation on a known lesion region rather than locating the lesion from the full mammogram. The effect of this lesion-centered setup on the interpretation of the results is discussed in Section~\ref{subsec:limitations}.

\subsubsection{Canonical patch construction}
\label{subsubsec:canonical}

Masses in CBIS-DDSM vary in size and orientation. To reduce this variation, each lesion-centered patch was normalized before training. The largest connected part of the binary mask was kept. Principal Component Analysis (PCA) was then applied to its pixel coordinates to estimate the lesion center, main orientation, and long-axis length. The isotropic scale factor is
\begin{equation}
s = \frac{L_t}{L},
\label{eq:scale}
\end{equation}
where $s$ is the dimensionless isotropic scale factor applied equally in the horizontal and vertical directions; $L$ is the lesion long-axis length estimated by PCA in the current patch; and $L_t$ is the fixed target long-axis length used for the canonical representation. A value $s>1$ therefore enlarges a lesion smaller than the target, and $s<1$ reduces a larger one. The same rotation-and-scaling transform, based on the estimated center, orientation, and $s$, was applied to both the image and mask so they remained aligned. The local ROI, the global contextual image, and the target mask were extracted from the transformed data.

\begin{algorithm}[!htbp]
\caption{Preprocessing and construction of the local and global input views.}
\label{alg:preprocessing}
\begin{algorithmic}[1]
\REQUIRE Mammogram $I$; binary lesion mask $M$; target canonical long-axis length $L_t$
\ENSURE Local input $\mathbf{X}_L$ of size $256\times256\times3$; global input $\mathbf{X}_G$ of size $512\times512\times3$; target mask $M_{\text{out}}$ of size $256\times256$
\STATE Clip $I$ at its lower and upper intensity percentiles and rescale to $[0,1]$, yielding $I_w$ \COMMENT{Section~\ref{subsubsec:intensity}}
\STATE Apply the Gabor filter to $I_w$ and normalize the response to $[0,1]$, yielding $I_g$
\STATE Binarize $M$ and retain its largest connected component
\STATE Apply PCA to the coordinates of the retained lesion pixels to obtain the lesion center, the dominant orientation, and the long-axis length $L$
\STATE Compute the scale factor $s \gets L_t/L$ using Eq.~\eqref{eq:scale}
\STATE Construct a rigid-plus-scale transform from the center, orientation, and $s$, and apply it jointly to $I_w$, $I_g$, and $M$ \COMMENT{joint application preserves image-mask alignment}
\STATE Crop the lesion-centered ROI and resize to $256\times256$; resize the corresponding mask region identically to obtain $M_{\text{out}}$
\STATE Crop the wider contextual view about the same center and resize to $512\times512$
\STATE Stack the channels according to Eq.~\eqref{eq:representation} to form $\mathbf{X}_L$ and $\mathbf{X}_G$
\RETURN $\mathbf{X}_L$, $\mathbf{X}_G$, $M_{\text{out}}$
\end{algorithmic}
\end{algorithm}

\subsection{Network Architecture}
\label{subsec:model_architecture}

\subsubsection{Overview}
\label{subsubsec:overview}

\begin{figure}[!htbp]
    \centering
    \safeincludegraphics[width=0.8\textwidth]{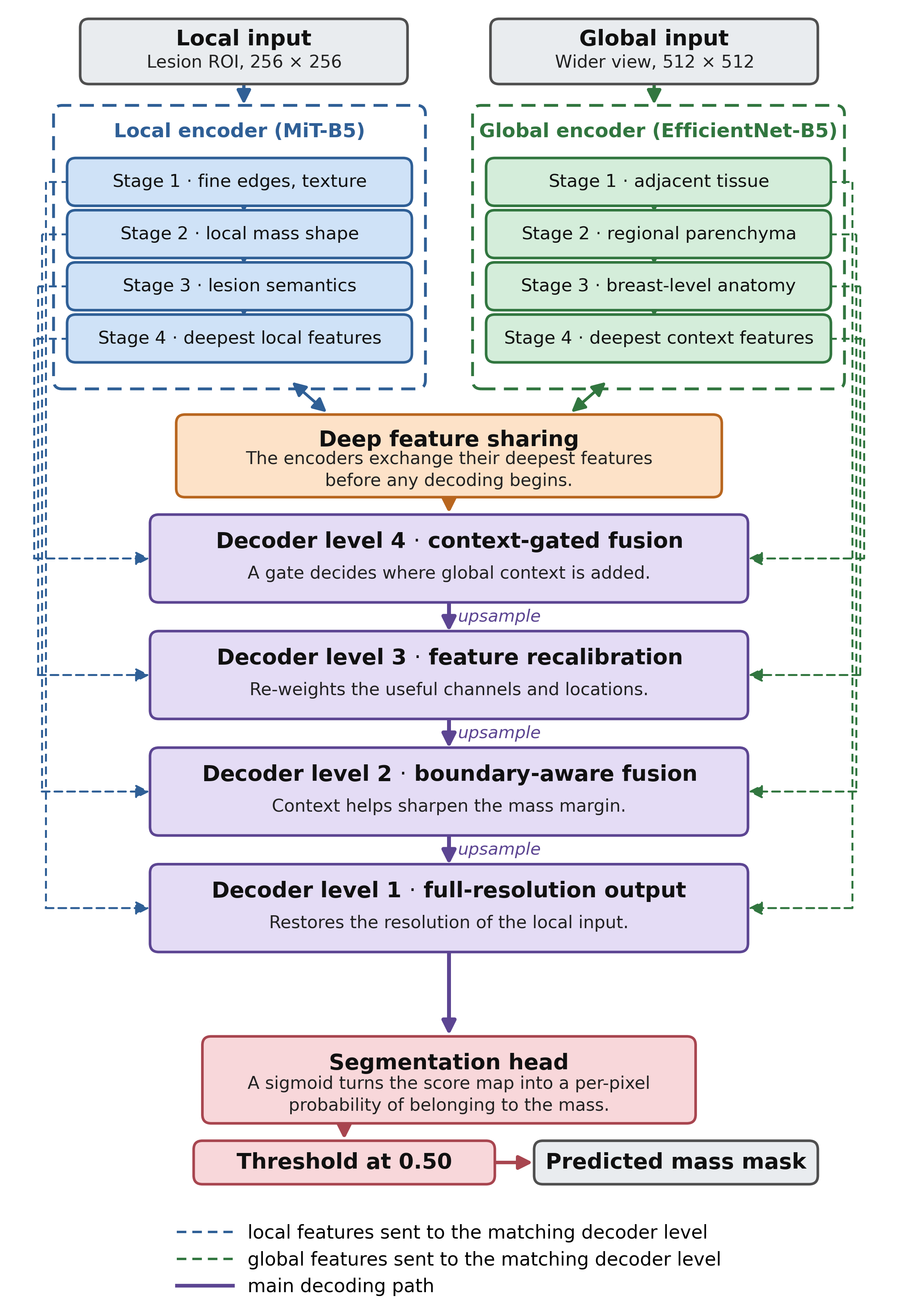}
    \caption{DualMiT-Net architecture. The local and global branches are shown in blue and green, respectively, and their features are combined progressively in the decoder.}
    \label{fig:model_arch}
\end{figure}

DualMiT-Net receives two views of the same lesion. The local input $\mathbf{X}_L \in \mathbb{R}^{256\times256\times3}$ contains the mass and its immediate boundary, while the global input $\mathbf{X}_G \in \mathbb{R}^{512\times512\times3}$ contains a wider region of surrounding breast tissue. The local view is processed by a MiT-B5 encoder and the global view by an EfficientNet-B5 encoder. Each encoder produces features at four spatial scales. Information is shared between the branches at the two deepest encoder levels, after which a single decoder reconstructs the segmentation mask from coarse to fine resolution. During decoding, local features provide the main lesion representation and global features are added through a spatial gate that controls where the wider context is useful. Fig.~\ref{fig:model_arch} shows the complete architecture, and Algorithm~\ref{alg:forward} summarizes the forward pass.

\subsubsection{Dual-Branch Encoder}
\label{subsubsec:encoders}

The two inputs are first processed independently. The local branch uses MiT-B5 to learn lesion shape, texture, and boundary information, while the global branch uses EfficientNet-B5 to learn information from the surrounding breast tissue. Both encoders provide four feature levels for later fusion in the decoder:
\begin{equation}
\{\mathbf{F}^{(i)}_L\}_{i=1}^{S} = E_L(\mathbf{X}_L),
\label{eq:local_feats}
\end{equation}
\begin{equation}
\{\mathbf{F}^{(i)}_G\}_{i=1}^{S} = E_G(\mathbf{X}_G).
\label{eq:global_feats}
\end{equation}
Here, $E_L$ and $E_G$ denote the local and global encoders, respectively, and $S=4$ is the number of retained feature levels. $\mathbf{F}^{(i)}_L$ and $\mathbf{F}^{(i)}_G$ are the local and global feature maps at level $i$. Level $i=1$ has the highest spatial resolution and contains more fine image detail, while level $i=S$ has the lowest spatial resolution and contains more abstract information.

MiT-B5 was selected for the local branch because its self-attention can model relationships across the lesion ROI while its hierarchical design provides multi-scale features for segmentation~\cite{xie2021segformer}. EfficientNet-B5 was selected for the global branch because it can capture a wider image context with a comparatively efficient convolutional encoder~\cite{tan2019efficientnet}. The two encoders therefore serve different roles: the local branch emphasizes the lesion itself, while the global branch summarizes the tissue around it.

\subsubsection{Deep Feature Sharing}
\label{subsubsec:sharing}

After the two encoders extract their features, information is shared between the branches at the two deepest levels. Sharing is limited to these deeper levels because they contain higher-level information and have lower spatial resolution, making the two feature streams easier to align than at the earlier encoder stages.

Feature sharing is performed in both directions. Local-to-global sharing projects the deep local features into the global feature space. Global-to-local sharing projects the deep global features into the local feature space and adds them to the local representation before decoding. In the remainder of this paper, G2L refers to global-to-local feature sharing. A variant without G2L sharing is evaluated in the ablation study in Section~\ref{subsec:component_ablation}.

\subsubsection{Context-Gated Fusion Decoder}
\label{subsubsec:decoder}

The decoder reconstructs the segmentation mask from the deepest feature level to the highest-resolution level. At each stage, the decoder combines three sources of information: the output from the preceding decoder stage, the local encoder features at the corresponding level, and the global encoder features resampled to the same spatial size. The local features remain the main segmentation pathway, while the global features provide additional context.

Let $\mathbf{Z}^{(i)}$ denote the decoder feature map at level $i$ before global context is added. The global information is introduced in three steps. First, $1\times1$ convolutions project the decoder and global features into query, key, and value maps:
\begin{equation}
\mathbf{Q}^{(i)} = \phi_q(\mathbf{Z}^{(i)}),\quad
\mathbf{K}^{(i)} = \phi_k(\mathbf{F}^{(i)}_G),\quad
\mathbf{V}^{(i)} = \phi_v(\mathbf{F}^{(i)}_G).
\label{eq:projections}
\end{equation}
$\mathbf{Q}^{(i)}$ represents the current decoder features, while $\mathbf{K}^{(i)}$ and $\mathbf{V}^{(i)}$ represent the global contextual features. The functions $\phi_q$, $\phi_k$, and $\phi_v$ are learnable $1\times1$ convolutions that produce a common projection dimension $d$.

Second, a spatial gate measures how well the decoder and global features agree at each image location:
\begin{equation}
\mathbf{A}^{(i)} = \sigma\!\left(\frac{\sum_{c=1}^{d} \mathbf{Q}^{(i)}_c \odot \mathbf{K}^{(i)}_c}{\sqrt{d}}\right).
\label{eq:spatial_gate}
\end{equation}
Here, $\mathbf{A}^{(i)}$ is a single-channel gate with values between 0 and 1, $c$ indexes the projected feature channels, $\odot$ denotes element-wise multiplication, and $\sigma$ is the sigmoid function. A larger gate value allows more global information to enter at that location, while a smaller value reduces its contribution.

Third, the gated global features are added to the decoder representation:
\begin{equation}
\tilde{\mathbf{Z}}^{(i)} = \mathbf{Z}^{(i)} + \gamma_i\left(\mathbf{A}^{(i)} \odot \mathbf{V}^{(i)}\right).
\label{eq:fusion_update}
\end{equation}
$\tilde{\mathbf{Z}}^{(i)}$ is the decoder feature map after contextual fusion, and $\gamma_i$ is a learnable scalar that controls the amount of global information added at level $i$. This residual form keeps the local decoder representation available while allowing the model to use wider breast context where it is helpful. Because the gate is calculated independently at each spatial position rather than through a full pairwise attention matrix, its cost grows linearly with the number of image positions.

After contextual fusion, each decoder level applies a concurrent Spatial and Channel Squeeze-and-Excitation (scSE) module~\cite{roy2018concurrent}. The channel component emphasizes useful feature channels, and the spatial component emphasizes useful image locations. Their outputs are combined to recalibrate the decoder features. A variant without scSE is evaluated in Section~\ref{subsec:component_ablation}.

\subsubsection{Segmentation Head}
\label{subsubsec:head}

After the final decoder stage, a $1\times1$ convolution converts the fused features into a single-channel score map:
\begin{equation}
\hat{\mathbf{Y}} = H(\tilde{\mathbf{Z}}^{(1)}),
\label{eq:seg_head}
\end{equation}
where $H$ denotes the segmentation head and $\hat{\mathbf{Y}}$ is the unbounded score map. A sigmoid function converts these scores to pixel-wise mass probabilities:
\begin{equation}
\mathbf{P} = \sigma(\hat{\mathbf{Y}}),
\label{eq:sigmoid}
\end{equation}
where $\mathbf{P}$ is the predicted probability map. The final binary mask is obtained using a fixed threshold of $\tau=0.50$:
\begin{equation}
\hat{\mathbf{M}} = \mathbf{1}\left[\mathbf{P} \geq \tau\right],
\label{eq:threshold}
\end{equation}
where $\hat{\mathbf{M}}$ denotes the predicted mass mask and $\mathbf{1}[\cdot]$ is the element-wise indicator function. The threshold is fixed for all experiments, as described in Section~\ref{subsubsec:thresholding}.

\begin{algorithm}[!htbp]
\caption{Forward pass of DualMiT-Net from the local and global inputs to the predicted mass mask.}
\label{alg:forward}
\begin{algorithmic}[1]
\REQUIRE Local input $\mathbf{X}_L$; global input $\mathbf{X}_G$; threshold $\tau=0.50$
\ENSURE Predicted binary mass mask $\hat{\mathbf{M}}$
\STATE $\{\mathbf{F}^{(i)}_L\}_{i=1}^{S} \gets E_L(\mathbf{X}_L)$ \COMMENT{Eq.~\eqref{eq:local_feats}}
\STATE $\{\mathbf{F}^{(i)}_G\}_{i=1}^{S} \gets E_G(\mathbf{X}_G)$ \COMMENT{Eq.~\eqref{eq:global_feats}}
\STATE Exchange features between branches at the two deepest levels \COMMENT{Section~\ref{subsubsec:sharing}}
\STATE $\mathbf{Z}^{(S)} \gets \mathbf{F}^{(S)}_L$
\FOR{$i = S$ \TO $1$}
    \STATE Resample $\mathbf{F}^{(i)}_G$ to the spatial size of $\mathbf{Z}^{(i)}$
    \STATE Compute $\mathbf{Q}^{(i)},\mathbf{K}^{(i)},\mathbf{V}^{(i)}$ using Eq.~\eqref{eq:projections}
    \STATE Compute the spatial gate $\mathbf{A}^{(i)}$ using Eq.~\eqref{eq:spatial_gate}
    \STATE Compute the fused features $\tilde{\mathbf{Z}}^{(i)}$ using Eq.~\eqref{eq:fusion_update}
    \STATE Apply feature recalibration to $\tilde{\mathbf{Z}}^{(i)}$
    \IF{$i > 1$}
        \STATE Upsample $\tilde{\mathbf{Z}}^{(i)}$ and concatenate with $\mathbf{F}^{(i-1)}_L$ to form $\mathbf{Z}^{(i-1)}$
    \ENDIF
\ENDFOR
\STATE $\hat{\mathbf{Y}} \gets H(\tilde{\mathbf{Z}}^{(1)})$ \COMMENT{Eq.~\eqref{eq:seg_head}}
\STATE $\mathbf{P} \gets \sigma(\hat{\mathbf{Y}})$ \COMMENT{Eq.~\eqref{eq:sigmoid}}
\STATE $\hat{\mathbf{M}} \gets \mathbf{1}[\mathbf{P} \geq \tau]$ \COMMENT{Eq.~\eqref{eq:threshold}}
\RETURN $\hat{\mathbf{M}}$
\end{algorithmic}
\end{algorithm}

\subsection{Loss Function}
\label{subsec:loss_function}

The network was optimized with a weighted sum of Binary Cross-Entropy (BCE) loss and soft Dice loss. The BCE term provides pixel-wise supervision, whereas the Dice term directly rewards overlap between the predicted and ground-truth masks:
\begin{equation}
\mathcal{L}_{\mathrm{total}} = \lambda_{\mathrm{BCE}}\, \mathcal{L}_{\mathrm{BCE}} + \lambda_{\mathrm{Dice}}\, \mathcal{L}_{\mathrm{Dice}},
\label{eq:total_loss}
\end{equation}
where $\mathcal{L}_{\mathrm{total}}$ denotes the total training objective; $\mathcal{L}_{\mathrm{BCE}}$ denotes the BCE loss between the predicted probability map $\mathbf{P}$ of Eq.~\eqref{eq:sigmoid} and the binary target mask $M_{\text{out}}$; $\mathcal{L}_{\mathrm{Dice}}$ denotes the soft Dice loss, a differentiable relaxation of the Dice overlap of Eq.~\eqref{eq:dice} in which the binary prediction is replaced by $\mathbf{P}$; and $\lambda_{\mathrm{BCE}}$ and $\lambda_{\mathrm{Dice}}$ are scalar coefficients controlling the relative contribution of the two losses, fixed at $\lambda_{\mathrm{BCE}}=0.3$ and $\lambda_{\mathrm{Dice}}=0.7$. Thus, the Dice term contributes 70\% of the weighted objective and the BCE term contributes 30\%.

The two terms play complementary roles. The BCE term is evaluated independently at each pixel and therefore supplies a dense gradient from the first epoch, which stabilizes early optimization. The Dice term evaluates the predicted mask as a whole and is insensitive to the proportion of the image occupied by background~\cite{milletari2016v,sudre2017generalised}. The weighting favors the Dice term because regional overlap is the primary criterion by which the model is assessed, while the smaller BCE weight is retained for its stabilizing effect.

\subsection{Evaluation Metrics}
\label{subsec:evaluation_metrics}

Segmentation quality was assessed using two families of metrics, the first quantifying regional agreement and the second quantifying the distance between predicted and ground-truth boundaries~\cite{muller2022towards,maier2024metrics}. For a single test image, a true positive (TP) is a mass pixel correctly predicted as mass, a false positive (FP) is a background pixel incorrectly predicted as mass, a false negative (FN) is a mass pixel missed by the model, and a true negative (TN) is a background pixel correctly predicted as background. The symbols TP, FP, FN, and TN in Eqs.~\eqref{eq:dice}-\eqref{eq:specificity} denote the corresponding pixel counts.

\subsubsection{Overlap metrics}
\label{subsubsec:overlap_metrics}

The Dice coefficient quantifies agreement between the predicted and ground-truth masks:
\begin{equation}
\mathrm{Dice} = \frac{2\,\mathrm{TP}}{2\,\mathrm{TP} + \mathrm{FP} + \mathrm{FN}},
\label{eq:dice}
\end{equation}
where TP, FP, and FN are as defined above. The Dice coefficient equals one for exact agreement and zero when the two masks do not intersect.

The Intersection over Union (IoU) quantifies the same agreement relative to the union of the two masks:
\begin{equation}
\mathrm{IoU} = \frac{\mathrm{TP}}{\mathrm{TP} + \mathrm{FP} + \mathrm{FN}}.
\label{eq:iou}
\end{equation}
Dice and IoU measure the same type of mask overlap, but IoU gives a lower numerical score for the same prediction and is therefore a stricter measure. 

\subsubsection{Pixel-classification metrics}
\label{subsubsec:pixel_metrics}

Three additional metrics help show how the remaining segmentation errors occur. Precision is the proportion of predicted mass pixels that are correct so that low values indicate over-segmentation into adjacent tissue:
\begin{equation}
\mathrm{Precision} = \frac{\mathrm{TP}}{\mathrm{TP} + \mathrm{FP}},
\label{eq:precision}
\end{equation}
Recall, equivalently sensitivity, is the proportion of ground-truth mass pixels recovered so that low values indicate under-segmentation:
\begin{equation}
\mathrm{Recall} = \frac{\mathrm{TP}}{\mathrm{TP} + \mathrm{FN}},
\label{eq:recall}
\end{equation}
Specificity is the proportion of background pixels correctly left unlabeled:
\begin{equation}
\mathrm{Specificity} = \frac{\mathrm{TN}}{\mathrm{TN} + \mathrm{FP}}.
\label{eq:specificity}
\end{equation}
Precision and recall are interpreted together because they show whether the model tends to over-segment or under-segment the mass.
Pixel accuracy is the proportion of all correctly classified pixels:
\begin{equation}
\mathrm{Accuracy} =
\frac{\mathrm{TP}+\mathrm{TN}}
{\mathrm{TP}+\mathrm{TN}+\mathrm{FP}+\mathrm{FN}}.
\label{eq:accuracy}
\end{equation}
Pixel accuracy is reported for completeness. Because background pixels usually occupy a large portion of the lesion-centered patch, a high accuracy value does not necessarily indicate accurate mass segmentation. For this reason, Dice and IoU are treated as the primary overlap metrics.

\subsubsection{Boundary metric}
\label{subsubsec:boundary_metric}

Overlap metrics are dominated by the interior of the mass and are comparatively insensitive to the accuracy of the margin, which is of primary clinical interest. The 95th-percentile Hausdorff Distance (HD95) is therefore also reported:
\begin{equation}
\mathrm{HD95} = \max\left\{
\underset{a \in  A}{P_{95}} \; \min_{b \in  B} \lVert a-b \rVert, \;\;
\underset{b \in  B}{P_{95}} \; \min_{a \in  A} \lVert a-b \rVert
\right\},
\label{eq:hd95}
\end{equation}
where $ A$ denotes the set of pixels lying on the predicted mask boundary; $ B$ denotes the set of pixels lying on the ground-truth mask boundary; $a$ and $b$ denote individual boundary pixels of $ A$ and $ B$ respectively; $\lVert a-b \rVert$ denotes the Euclidean distance between two such pixels; $\min_{b \in  B} \lVert a-b \rVert$ denotes the distance from a predicted boundary pixel to the nearest ground-truth boundary pixel; $P_{95}$ denotes the 95th percentile of the set of distances over which it is taken; and $\max\{\cdot,\cdot\}$ denotes selection of the larger of the two directed percentile distances. Use of the 95th percentile rather than the maximum confers robustness to isolated outlier pixels, and lower values indicate closer boundary agreement. HD95 is expressed in pixels of the $256\times256$ local grid. Because the canonical transform of Section~\ref{subsubsec:canonical} rescales each lesion, these distances are not convertible to millimeters and are comparable only between models evaluated on the same processed data, as is the case throughout this study.

\subsubsection{Inference and metric aggregation}
\label{subsubsec:thresholding}

All metrics defined above operate on binary masks, so the probability map $\mathbf{P}$ is thresholded before evaluation according to Eq.~\eqref{eq:threshold}. The threshold was fixed at $\tau = 0.50$ for every model, input representation, and seed, and was not tuned on the test set. Since $\tau$ trades precision against recall, per-model tuning on test data would confound the comparisons reported in Section~\ref{sec:results}. Metrics were computed independently for each of the 92 test images and then averaged, so that each case contributes equally irrespective of lesion size.

\subsection{Experimental Protocol}
\label{subsec:protocol}

\subsubsection{Training configuration}
\label{subsubsec:training_config}

All models were trained for a maximum of 150 epochs with a batch size of four, using Adam with decoupled weight decay (AdamW)~\cite{loshchilov2019decoupled}. The learning rate was $5\times10^{-5}$ and the weight decay $1\times10^{-4}$. Gradients were clipped to a maximum norm of 1.0. The learning rate was halved whenever validation Dice failed to improve, and training was terminated early if validation Dice had not improved over 20 consecutive epochs.

An exponential moving average (EMA) of the model parameters was maintained alongside the directly optimized parameters~\cite{polyak1992acceleration}. Following each optimizer step, the averaged parameters were updated according to
\begin{equation}
\theta_{\mathrm{EMA}} \leftarrow \beta\,\theta_{\mathrm{EMA}} + (1-\beta)\,\theta,
\label{eq:ema}
\end{equation}
where $\theta$ denotes the model parameters produced directly by the optimizer, referred to below as the base weights; $\theta_{\mathrm{EMA}}$ denotes the exponentially averaged parameters, referred to below as the EMA weights; and $\beta$ denotes the decay factor, set to $0.999$. Because $\beta$ is close to unity, $\theta_{\mathrm{EMA}}$ evolves slowly and follows a smoothed trajectory through parameter space, which frequently generalizes marginally better than the final iterate. Both parameter sets were evaluated and are reported alongside one another. Training is summarized in Algorithm~\ref{alg:training}. All experiments were implemented in PyTorch and executed on a workstation equipped with an Intel Core Ultra 9 processor and an NVIDIA GeForce RTX 4090 graphics processing unit.

\begin{algorithm}[!htbp]
\caption{Training procedure with EMA weights and early stopping.}
\label{alg:training}
\begin{algorithmic}[1]
\REQUIRE Training set $\mathcal{D}_{\text{train}}$; validation set $\mathcal{D}_{\text{val}}$; maximum epochs $E_{\max}$; patience $P$; EMA decay $\beta$; loss weights $\lambda_{\mathrm{BCE}}$, $\lambda_{\mathrm{Dice}}$
\ENSURE Best base weights $\theta^*$; best EMA weights $\theta^*_{\mathrm{EMA}}$
\STATE Initialize $\theta$ from ImageNet-pretrained encoder weights
\STATE $\theta_{\mathrm{EMA}} \gets \theta$
\FOR{epoch $e = 1$ \TO $E_{\max}$}
    \FORALL{mini-batches $(\mathbf{X}_L, \mathbf{X}_G, M_{\text{out}}) \in \mathcal{D}_{\text{train}}$}
        \STATE Evaluate Algorithm~\ref{alg:forward} as far as the probability map $\mathbf{P}$
        \STATE Compute $\mathcal{L}_{\mathrm{total}}$ from $\mathbf{P}$ and $M_{\text{out}}$ using Eq.~\eqref{eq:total_loss}
        \STATE Back-propagate, clip gradients to norm 1.0, and update $\theta$ with AdamW
        \STATE Update $\theta_{\mathrm{EMA}}$ using Eq.~\eqref{eq:ema}
    \ENDFOR
    \STATE Evaluate validation Dice separately for $\theta$ and $\theta_{\mathrm{EMA}}$
    \STATE Retain the best-performing checkpoint of each
    \STATE Halve the learning rate if validation Dice has not improved
    \IF{validation Dice has not improved over $P$ consecutive epochs}
        \STATE \textbf{break}
    \ENDIF
\ENDFOR
\RETURN $\theta^*$, $\theta^*_{\mathrm{EMA}}$
\end{algorithmic}
\end{algorithm}

\subsubsection{Statistical analysis}
\label{subsubsec:statistics}

Every reported configuration was trained three times using seeds 42, 43, and 44. Results are reported as the mean and standard deviation across the three runs. Repeating the experiments helps separate the effect of a model change from normal variation caused by random initialization.

Two additional analyses were performed using the per-image scores averaged across the three seeds. This produced one score for each of the 92 test images. To calculate 95\% bootstrap confidence intervals, the 92 cases were resampled with replacement 10,000 times and the mean was recalculated for each sample. The 2.5th and 97.5th percentiles were used as the lower and upper limits of the interval. These intervals describe variation due to the test cases; variation between training runs is reported separately with the across-seed standard deviation. Model variants were also compared with the paired Wilcoxon signed-rank test using scores from the same 92 images. Because the same images were used for both models, the comparison directly measures the paired difference for each case. A difference was considered statistically significant when $p<0.05$.

\section{Results}
\label{sec:results}

The results are presented in five parts. First, the overall segmentation performance of DualMiT-Net is reported. We then examine the input representation and the contribution of the main architectural components. Qualitative examples are used to show successful and difficult segmentation cases. Finally, DualMiT-Net is compared with standard segmentation baselines and previously published results on CBIS-DDSM.

\subsection{Segmentation Performance on CBIS-DDSM}
\label{subsec:main_results}

Table~\ref{tab:three_seed_candidates} presents the test results for the full DualMiT-Net and the two ablation variants. Each model was trained with three seeds, and the results are reported as mean $\pm$ standard deviation across the three runs. Results are shown for both the base weights and the EMA weights. 
The full DualMiT-Net with EMA weights achieved the highest overlap scores, with a Dice coefficient of $0.9375 \pm 0.0011$ and an Intersection over Union (IoU) of $0.8834 \pm 0.0020$. The corresponding base model achieved a Dice coefficient of $0.9369 \pm 0.0012$ and an IoU of $0.8824 \pm 0.0020$. The small standard deviations across the three runs indicate that the model produced similar results across different seeds.

The precision and recall values provide additional information about the predicted mass boundaries. For the full EMA model, precision was $0.9282 \pm 0.0052$ and recall was $0.9482 \pm 0.0037$. The slightly higher recall indicates that the model recovered most of the annotated lesion area while including a small amount of additional tissue outside the expert boundary. This pattern is more consistent with mild over-segmentation than with large missed lesion regions. The HD95 values were below $7$ pixels for all configurations, showing that the predicted boundaries remained close to the expert boundaries. Overall, the complete DualMiT-Net provided the strongest overlap performance among the three architectural configurations.

\begin{table*}[!htbp]
\centering
\caption{Segmentation performance of DualMiT-Net and the two ablation variants.}
\label{tab:three_seed_candidates}
\resizebox{\textwidth}{!}{%
\begin{tabular}{lllllllll}
\toprule
Architecture & Weights & Dice & IoU & Precision & Recall & Specificity & Accuracy & HD95 $\downarrow$ \\
\midrule
Full DualMiT-Net & Base & $0.9369 \pm 0.0012$ & $0.8824 \pm 0.0020$ & $0.9333 \pm 0.0024$ & $0.9418 \pm 0.0040$ & $0.9102 \pm 0.0039$ & $0.9278 \pm 0.0012$ & $6.85 \pm 0.25$ \\
Full DualMiT-Net & EMA & $\mathbf{0.9375 \pm 0.0011}$ & $\mathbf{0.8834 \pm 0.0020}$ & $0.9282 \pm 0.0052$ & $0.9482 \pm 0.0037$ & $0.9022 \pm 0.0080$ & $0.9280 \pm 0.0015$ & $6.88 \pm 0.23$ \\
Without scSE & Base & $0.9368 \pm 0.0011$ & $0.8822 \pm 0.0019$ & $0.9368 \pm 0.0035$ & $0.9380 \pm 0.0050$ & $0.9158 \pm 0.0052$ & $0.9280 \pm 0.0010$ & $6.82 \pm 0.14$ \\
Without scSE & EMA & $0.9372 \pm 0.0002$ & $0.8829 \pm 0.0004$ & $0.9238 \pm 0.0012$ & $0.9522 \pm 0.0014$ & $0.8953 \pm 0.0018$ & $0.9273 \pm 0.0002$ & $6.95 \pm 0.09$ \\
Without G2L sharing & Base & $0.9363 \pm 0.0003$ & $0.8814 \pm 0.0006$ & $0.9267 \pm 0.0065$ & $0.9474 \pm 0.0065$ & $0.8998 \pm 0.0103$ & $0.9266 \pm 0.0007$ & $6.97 \pm 0.17$ \\
Without G2L sharing & EMA & $0.9371 \pm 0.0006$ & $0.8829 \pm 0.0011$ & $0.9273 \pm 0.0026$ & $0.9484 \pm 0.0018$ & $0.9008 \pm 0.0038$ & $0.9275 \pm 0.0008$ & $6.93 \pm 0.06$ \\
\bottomrule
\end{tabular}}
\end{table*}

\subsection{Input Representation Ablation}
\label{subsec:input_ablation}

Four input representations were evaluated using the same DualMiT-Net architecture: windowing alone, windowing with CLAHE, windowing with a Gabor texture response, and windowing with both CLAHE and Gabor filtering. Table~\ref{tab:input_ablation} summarizes the results.

Windowing combined with the Gabor response produced the highest Dice and IoU values for both base and EMA weights. With EMA weights, Dice increased from $0.9363$ with windowing alone to $0.9376$ with windowing and Gabor filtering, while IoU increased from $0.8815$ to $0.8836$. With the base weights, the same representation achieved a Dice coefficient of $0.9372$ and an IoU of $0.8829$.  
The window-and-Gabor representation also produced the lowest HD95 for the base model at $6.70$ pixels, showing that the improvement was observed in both lesion overlap and boundary agreement. 
Adding CLAHE did not provide a similar improvement. Windowing with CLAHE performed nearly the same as windowing alone, while combining CLAHE and Gabor filtering resulted in slightly lower Dice and IoU values. Based on these results, the window-and-Gabor representation was selected for the final model.

\begin{table*}[!htbp]
\centering
\caption{Comparison of the four input representations.}
\label{tab:input_ablation}
\resizebox{\textwidth}{!}{%
\begin{tabular}{lllllll}
\toprule
Input representation & Base Dice & Base IoU & Base HD95 & EMA Dice & EMA IoU & EMA HD95 \\
\midrule
Window only & $0.9364 \pm 0.0278$ & $0.8817 \pm 0.0480$ & $6.88 \pm 3.70$ & $0.9363 \pm 0.0277$ & $0.8815 \pm 0.0478$ & $7.00 \pm 3.93$ \\
Window + CLAHE & $0.9361 \pm 0.0268$ & $0.8811 \pm 0.0465$ & $7.09 \pm 4.13$ & $0.9363 \pm 0.0266$ & $0.8814 \pm 0.0462$ & $7.21 \pm 4.20$ \\
Window + Gabor & $\mathbf{0.9372 \pm 0.0248}$ & $\mathbf{0.8829 \pm 0.0433}$ & $\mathbf{6.70 \pm 2.90}$ & $\mathbf{0.9376 \pm 0.0250}$ & $\mathbf{0.8836 \pm 0.0438}$ & $\mathbf{6.85 \pm 3.43}$ \\
Window + CLAHE + Gabor & $0.9356 \pm 0.0276$ & $0.8802 \pm 0.0477$ & $7.01 \pm 3.97$ & $0.9356 \pm 0.0272$ & $0.8802 \pm 0.0470$ & $7.14 \pm 3.96$ \\
\bottomrule
\end{tabular}%
}
\end{table*}

\subsection{Component Ablation}
\label{subsec:component_ablation}

The ablation study examines how the performance of DualMiT-Net changes when individual refinement components are removed. The first variant removes the scSE feature-recalibration module. The second removes global-to-local (G2L) feature sharing while keeping the local-to-global direction. Both variants retain the dual-view encoders and the main decoder structure.

Table~\ref{tab:three_seed_candidates} shows that all three configurations achieved strong segmentation performance. The complete DualMiT-Net achieved a Dice coefficient of $0.9375$ and an IoU of $0.8834$ with EMA weights. The corresponding Dice scores were $0.9372$ without scSE and $0.9371$ without G2L sharing. 
The small changes after removing one component show that the overall dual-view architecture remains effective even when one refinement mechanism is omitted. The complete configuration preserved the strongest overall overlap performance and was therefore used as the final model.

\subsubsection{Bootstrap Confidence Intervals}
\label{subsubsec:bootstrap}

Table~\ref{tab:bootstrap_ci} presents the seed-averaged metrics together with 95\% bootstrap confidence intervals calculated from the 92 test images. 
For the full model with EMA weights, the mean Dice coefficient was $0.9375$, with a 95\% confidence interval of $[0.9342, 0.9406]$. The corresponding IoU was $0.8834$, with an interval of $[0.8777, 0.8889]$. Similar intervals were observed for the two ablation variants. 
The confidence intervals overlap because the three configurations produce similar predictions for many test images. This is consistent with the fact that each ablation retains most of the DualMiT-Net architecture. The complete configuration nevertheless maintained the highest mean Dice and IoU.

%\vspace{-15pt}

\begin{table*}[!htbp]
\centering
\caption{Seed-averaged results with 95\% bootstrap confidence intervals.}
\label{tab:bootstrap_ci}
\resizebox{\textwidth}{!}{%
\begin{tabular}{lllllllll}
\toprule
Architecture & Weights & Dice & IoU & Precision & Recall & Specificity & Accuracy & HD95 $\downarrow$ \\
\midrule
Full DualMiT-Net & Base & 0.9369 [0.9336, 0.9401] & 0.8824 [0.8767, 0.8879] & 0.9333 [0.9280, 0.9383] & 0.9418 [0.9364, 0.9467] & 0.9102 [0.9029, 0.9168] & 0.9278 [0.9242, 0.9312] & 6.85 [6.42, 7.33] \\
Full DualMiT-Net & EMA & 0.9375 [0.9342, 0.9406] & 0.8834 [0.8777, 0.8889] & 0.9282 [0.9212, 0.9344] & 0.9482 [0.9436, 0.9529] & 0.9022 [0.8924, 0.9111] & 0.9280 [0.9242, 0.9316] & 6.88 [6.45, 7.39] \\
Without G2L sharing & Base & 0.9363 [0.9331, 0.9393] & 0.8814 [0.8758, 0.8867] & 0.9267 [0.9194, 0.9345] & 0.9474 [0.9400, 0.9538] & 0.8998 [0.8892, 0.9115] & 0.9266 [0.9230, 0.9300] & 6.97 [6.57, 7.42] \\
Without G2L sharing & EMA & 0.9371 [0.9339, 0.9402] & 0.8829 [0.8772, 0.8883] & 0.9273 [0.9219, 0.9326] & 0.9484 [0.9445, 0.9522] & 0.9008 [0.8936, 0.9076] & 0.9275 [0.9239, 0.9310] & 6.93 [6.56, 7.33] \\
Without scSE & Base & 0.9368 [0.9334, 0.9400] & 0.8822 [0.8764, 0.8878] & 0.9368 [0.9310, 0.9422] & 0.9380 [0.9320, 0.9437] & 0.9158 [0.9081, 0.9230] & 0.9280 [0.9244, 0.9314] & 6.82 [6.44, 7.26] \\
Without scSE & EMA & 0.9372 [0.9340, 0.9402] & 0.8829 [0.8773, 0.8883] & 0.9238 [0.9187, 0.9287] & 0.9522 [0.9488, 0.9555] & 0.8953 [0.8889, 0.9016] & 0.9273 [0.9237, 0.9307] & 6.95 [6.54, 7.43] \\
\bottomrule
\end{tabular}}
\end{table*}

\begin{figure*}[!h]%[!htbp]
\centering

% First row
\begin{minipage}{0.48\textwidth}
\centering
\safeincludegraphics[width=\linewidth]{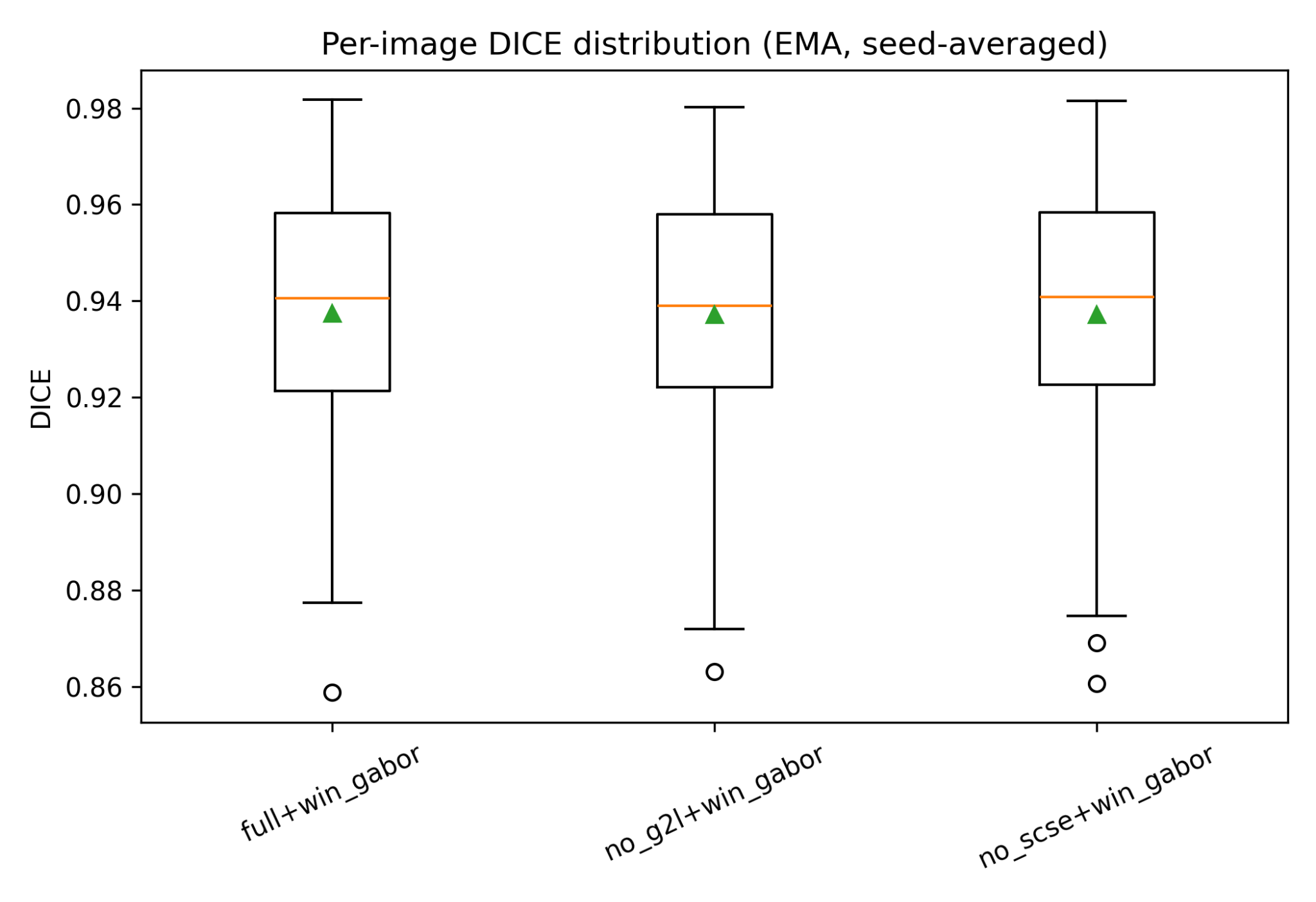}\\[0pt]
{\small (a) Dice}
\end{minipage}
\hfill
\begin{minipage}{0.48\textwidth}
\centering
\safeincludegraphics[width=\linewidth]{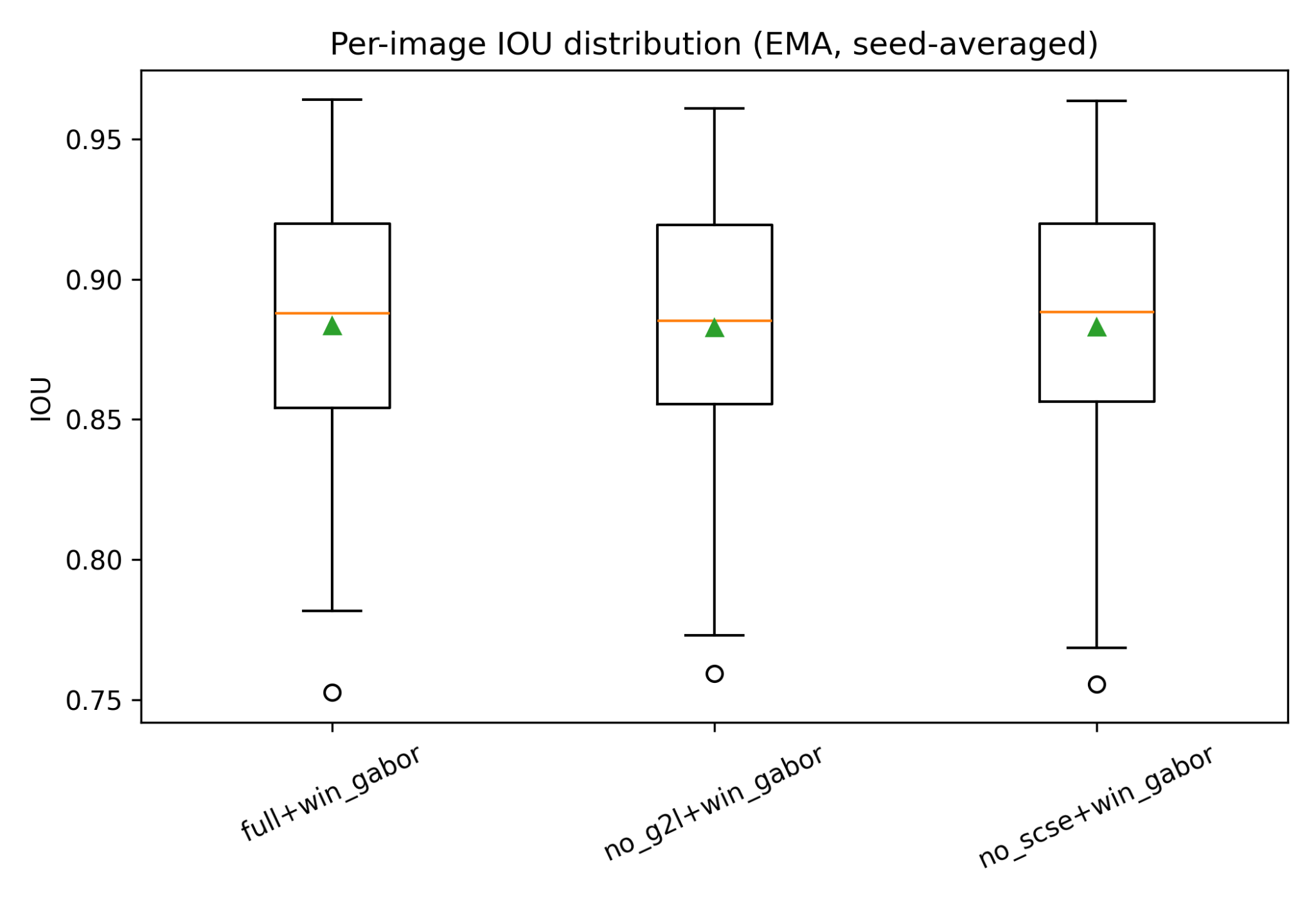}\\[0pt]
{\small (b) IoU}
\end{minipage}

%\vspace{0.5em}

% Second row
\begin{minipage}{0.48\textwidth}
\centering
\safeincludegraphics[width=\linewidth]{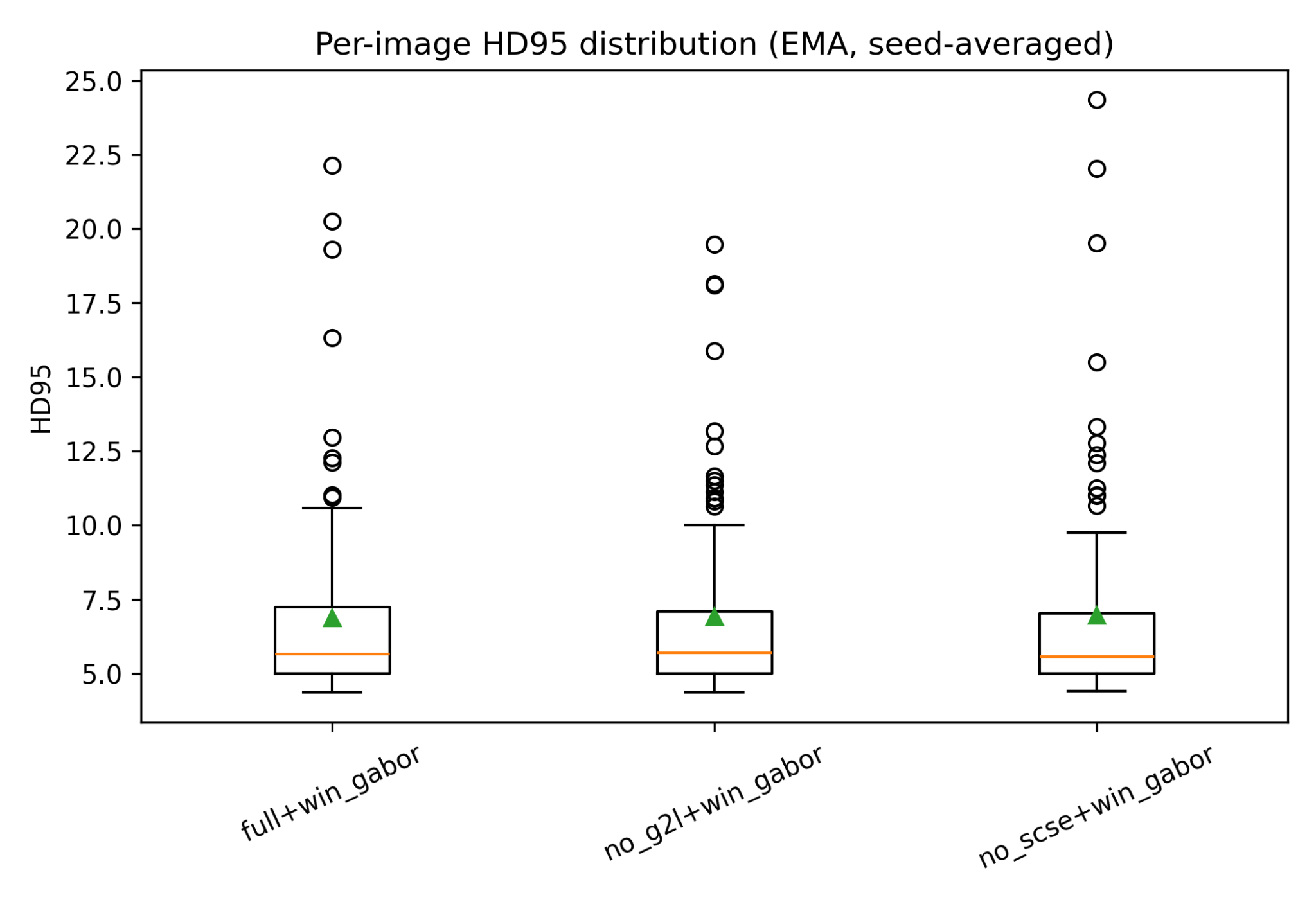}\\[0pt]
{\small (c) HD95}
\end{minipage}

\caption{Per-image Dice, IoU, and HD95 distributions for the three architectures using EMA weights.}
\label{fig:ema_distributions}
\end{figure*}

Fig.~\ref{fig:ema_distributions} shows the per-image Dice, IoU, and HD95 distributions for the same three configurations using EMA weights. The distributions are closely grouped, showing that the main segmentation behavior is preserved across the ablation variants. 
HD95 shows greater variation for difficult cases, with several larger values in the upper part of the distribution. These cases are examined further in Section~\ref{subsec:qualitative}.

%\vspace{-30pt}

\subsubsection{Paired Significance Tests}
\label{subsubsec:paired_tests}

Paired Wilcoxon signed-rank tests were used to examine the per-image differences between the complete DualMiT-Net and each ablation variant. Table~\ref{tab:wilcoxon} reports the mean paired difference, 95\% confidence interval, and $p$-value for Dice, IoU, and HD95.

The mean paired differences were in favor of the complete configuration for all six comparisons. The differences ranged from $0.0003$ to $0.0006$ for Dice and IoU and from $0.05$ to $0.07$ pixels for HD95. None of the comparisons reached statistical significance at $p<0.05$. 
These results show that removing either scSE or G2L sharing does not cause a large change in segmentation performance. This is consistent with the architecture, since the ablation variants still retain the local and global branches, the remaining feature-sharing pathway, and the decoder. The statistical results therefore describe the contribution of the individual refinement components rather than the effectiveness of DualMiT-Net as a whole.

\vspace{-10pt} 
\begin{table}[!htbp]
\centering
\caption{Paired Wilcoxon signed-rank tests for the two ablation comparisons.}
\label{tab:wilcoxon}
\resizebox{\columnwidth}{!}{%
\begin{tabular}{lllllc}
\toprule
Comparison & Metric & Mean difference & 95\% CI & $p$-value & Significance \\
\midrule
Full vs.\ without scSE & Dice & $0.0003$ & $[-0.0005, 0.0011]$ & $0.501$ & n.s. \\
Full vs.\ without G2L sharing & Dice & $0.0004$ & $[-0.0003, 0.0010]$ & $0.291$ & n.s. \\
Full vs.\ without scSE & IoU & $0.0005$ & $[-0.0009, 0.0019]$ & $0.513$ & n.s. \\
Full vs.\ without G2L sharing & IoU & $0.0006$ & $[-0.0005, 0.0017]$ & $0.300$ & n.s. \\
Full vs.\ without scSE & HD95 & $-0.07$ & $[-0.23, 0.08]$ & $0.507$ & n.s. \\
Full vs.\ without G2L sharing & HD95 & $-0.05$ & $[-0.19, 0.10]$ & $0.315$ & n.s. \\
\bottomrule
\end{tabular}}
\end{table}

%\vspace{-15pt}

\subsection{Qualitative Results and Failure Cases}
\label{subsec:qualitative}

\begin{figure*}[!t]%[!htbp]
  \centering
  \safeincludegraphics[width=0.8\textwidth]{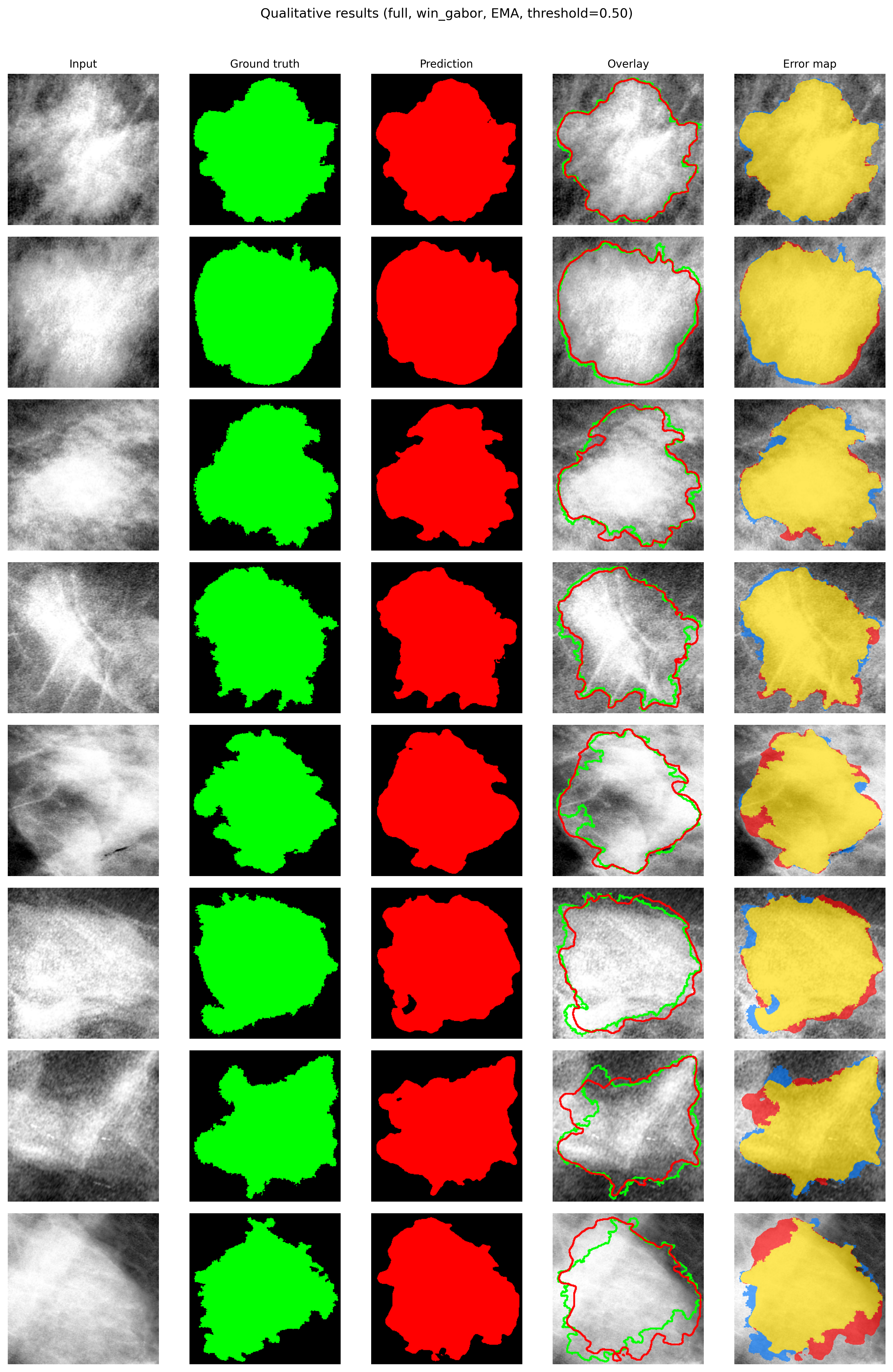}
  \caption{Representative segmentation results for DualMiT-Net.}
  \label{fig:qualitative_results}
\end{figure*}

Fig.~\ref{fig:qualitative_results} shows representative test cases for the full DualMiT-Net using EMA weights and the window-and-Gabor input. Each row represents one test case. The columns show the input ROI, expert mask, predicted mask, contour overlay, and error map.

In the contour overlay, the expert boundary is shown in green and the predicted boundary in red. In the error map, yellow represents pixels correctly identified as mass, red represents predicted mass pixels outside the expert mask, and blue represents mass pixels missed by the model. 
The predicted masks generally follow the expert mass boundaries closely. Most disagreement occurs near the lesion margin rather than across the main lesion area. Small red and blue regions appearing on opposite sides of some lesions indicate a slight shift in the predicted boundary rather than a large error in lesion location or size. 
These examples are consistent with the quantitative results. The high Dice and IoU values show strong overlap with the expert masks, while the remaining HD95 error is mainly associated with local differences along the lesion boundary.

\begin{figure*}[!t]%[!htbp]
  \centering
  \safeincludegraphics[width=0.8\textwidth]{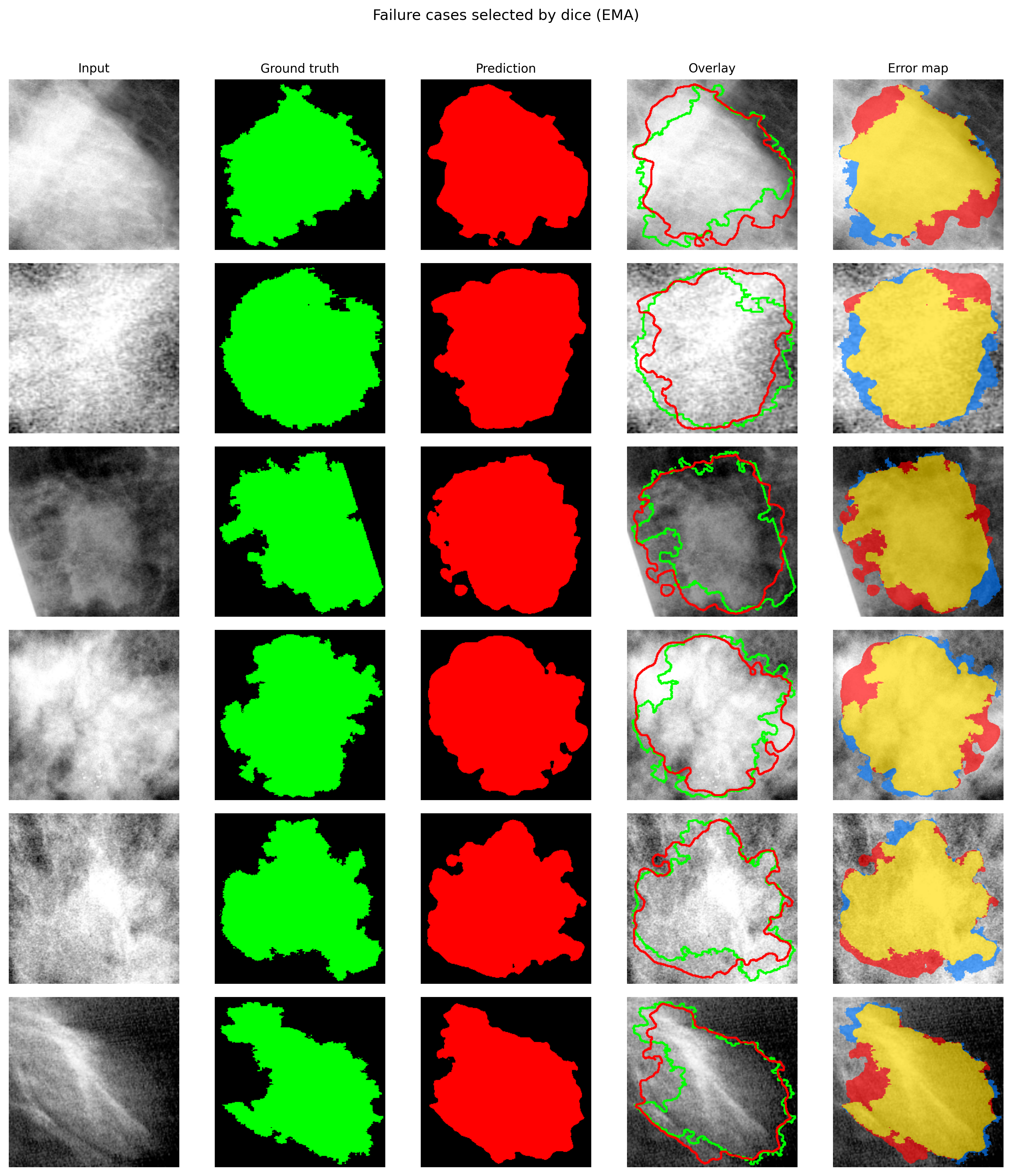}
  \caption{Test cases with the lowest Dice scores.}
  \label{fig:failure_cases}
\end{figure*}

Fig.~\ref{fig:failure_cases} shows the test cases with the lowest Dice scores. These cases commonly contain low contrast between the lesion and surrounding tissue, unclear or partly hidden mass boundaries, or background tissue with an appearance similar to the lesion. 
The red and blue error regions are wider in these cases, showing that most of the performance loss comes from disagreement along the mass boundary. The model generally identifies the correct lesion region, but the exact extent of the mass is less accurately reproduced. 
These difficult cases also help explain the upper values in the HD95 distribution shown in Fig.~\ref{fig:ema_distributions}(c). A relatively small local boundary error can produce a larger HD95 value even when most of the lesion remains correctly segmented.

\subsection{Comparison with Baseline Architectures}
\label{subsec:baseline_comparison}

Table~\ref{tab:unet_comparison} compares DualMiT-Net with six standard encoder-decoder segmentation models. All baseline models were trained using the same CBIS-DDSM data, preprocessing, loss function, optimizer, and prediction threshold.

Among the baseline models, U-Net with a ResNet-50 encoder achieved the highest Dice coefficient of $0.9025$ and IoU of $0.8240$. DualMiT-Net with EMA weights achieved a Dice coefficient of $0.9375$ and IoU of $0.8834$. This corresponds to absolute improvements of $0.0350$ in Dice and $0.0594$ in IoU over the strongest baseline. 
The six baseline models produced similar results, with their mean Dice scores ranging from $0.8941$ to $0.9025$. In comparison, both versions of DualMiT-Net achieved Dice scores above $0.936$. This clear separation shows that the proposed dual-view architecture provided a substantial improvement over the standard encoder-decoder models evaluated under the same experimental setting.

Precision and recall also improved with DualMiT-Net. The full EMA model achieved a recall of $0.9482$, compared with $0.9233$ for U-Net with ResNet-50, showing that DualMiT-Net recovered a larger portion of the expert-annotated lesion area. 
The reported standard deviations represent different sources of variation. For the baseline models, the spread is calculated across the 92 test images and reflects differences between test cases. For DualMiT-Net, it is calculated across three independent training runs and reflects variation between runs. These values should therefore not be compared directly.

\begin{table*}[!htbp]
\centering
\caption{Comparison of DualMiT-Net with standard segmentation baselines on CBIS-DDSM.}
\label{tab:unet_comparison}
\resizebox{\textwidth}{!}{%
\begin{tabular}{lcccc}
\toprule
Model & Dice & IoU & Precision & Recall \\
\midrule
U-Net (ResNet-50) & $0.9025 \pm 0.0339$ & $0.8240 \pm 0.0548$ & $0.8846 \pm 0.0524$ & $0.9233 \pm 0.0341$ \\
U-Net++ (ResNet-50) & $0.9023 \pm 0.0321$ & $0.8235 \pm 0.0523$ & $0.8806 \pm 0.0530$ & $0.9276 \pm 0.0337$ \\
U-Net++ (ResNet-34) & $0.9009 \pm 0.0315$ & $0.8211 \pm 0.0514$ & $0.8870 \pm 0.0527$ & $0.9174 \pm 0.0301$ \\
U-Net (ResNet-34) & $0.9005 \pm 0.0344$ & $0.8207 \pm 0.0555$ & $0.8772 \pm 0.0544$ & $0.9271 \pm 0.0290$ \\
FPN (ResNet-34) & $0.9002 \pm 0.0318$ & $0.8201 \pm 0.0514$ & $0.8770 \pm 0.0533$ & $0.9269 \pm 0.0274$ \\
PSPNet (ResNet-34) & $0.8941 \pm 0.0363$ & $0.8103 \pm 0.0582$ & $0.8787 \pm 0.0551$ & $0.9126 \pm 0.0402$ \\
\midrule
DualMiT-Net (Base) & $0.9369 \pm 0.0012$ & $0.8824 \pm 0.0020$ & $0.9333 \pm 0.0024$ & $0.9418 \pm 0.0040$ \\
DualMiT-Net (EMA) & $\mathbf{0.9375 \pm 0.0011}$ & $\mathbf{0.8834 \pm 0.0020}$ & $0.9282 \pm 0.0052$ & $\mathbf{0.9482 \pm 0.0037}$ \\
\bottomrule
\end{tabular}%
}
\end{table*}

\subsection{Comparison with Published Results}
\label{subsec:sota_comparison}

Table~\ref{tab:sota_comparison} compares DualMiT-Net with previously published breast mass segmentation results on CBIS-DDSM. Dice is reported for all methods in the table, while IoU is included only when it was reported in the original study.

Among the previous methods, Cross-view VAE reported the highest Dice coefficient at $0.9246$, with a Jaccard score of $0.8720$. Because the Jaccard index and IoU represent the same overlap measure, this value is reported in the IoU column. Connected-ResUNets reported a Dice coefficient of $0.8952$ and an IoU of $0.8002$. 
DualMiT-Net achieved a Dice coefficient of $0.9369$ and an IoU of $0.8824$ using the base weights. With EMA weights, Dice increased to $0.9375$ and IoU to $0.8834$. These are the highest values among the methods included in Table~\ref{tab:sota_comparison}. 
Compared with Cross-view VAE, the highest previously reported Dice value in the table, DualMiT-Net with EMA weights improved Dice by $0.0129$ and IoU by $0.0114$. These results show that DualMiT-Net achieved higher reported overlap values than the previous CBIS-DDSM methods included in the comparison.

\begin{table}[!htbp]
\centering
\caption{Comparison of breast mass segmentation methods on CBIS-DDSM.}
\label{tab:sota_comparison}
\begin{tabular}{lcc}
\toprule
Method & Dice & IoU \\
\midrule
YOLO-LOGO~\cite{su2022yolo} & 0.7450 & 0.6400 \\
AUNet~\cite{sun2020aunet} & 0.8180 & - \\
AM-MSP-cGAN~\cite{wang2020ammsp} & 0.8449 & - \\
ARF-Net~\cite{xu2022arfnet} & 0.8575 & - \\
MTLNet~\cite{hou2021mtlnet} & 0.8630 & - \\
Connected-ResUNets~\cite{baccouche2021connected} & 0.8952 & 0.8002 \\
Cross-view VAE~\cite{ma2024crossview} & 0.9246 & 0.8720 \\
\midrule
DualMiT-Net (Base) & 0.9369 & 0.8824 \\
DualMiT-Net (EMA) & \textbf{0.9375} & \textbf{0.8834} \\
\bottomrule
\end{tabular}
\end{table}

\section{Discussion}
\label{sec:discussion}

\subsection{Sources of Performance Gain}
\label{subsec:why_it_works}

The main performance gain of DualMiT-Net comes from its dual-view design. The model uses a local view to preserve lesion detail and a wider global view to retain information from the surrounding breast tissue. These complementary features are processed by separate encoders and combined during decoding. 
This design produced a clear improvement over the standard segmentation baselines. DualMiT-Net with EMA weights achieved a Dice coefficient of $0.9375$ and an IoU of $0.8834$, compared with $0.9025$ and $0.8240$ for the strongest baseline, U-Net with a ResNet-50 encoder. The improvement of $0.0350$ in Dice and $0.0594$ in IoU shows that the dual-view design provided a substantial gain over the conventional encoder-decoder models evaluated under the same setting.

The ablation study supports this interpretation. Removing scSE or G2L sharing produced only a small change because the main dual-view structure remained intact. The refinement components improve how local and global information is exchanged and recalibrated, while the core segmentation capability is provided by the complete local-global architecture. 
The input representation study also supported the final design. Windowing combined with the Gabor response produced the highest Dice and IoU among the four tested representations and also produced the lowest base HD95. This representation was therefore retained for the final model.

\subsection{Selection of the Final Configuration}
\label{subsec:why_full}

The complete DualMiT-Net was retained as the final configuration because it preserves the full local-global feature-fusion design and achieved the strongest overall overlap performance. 
G2L sharing allows contextual information from the global branch to refine the local representation before decoding. The scSE modules then recalibrate the fused decoder features in both the spatial and channel dimensions. These operations complement the two-view encoding strategy rather than acting as independent segmentation modules.

The precision and recall values also show how the complete model balances lesion recovery and boundary control. With EMA weights, the full model achieved a precision of $0.9282$ and recall of $0.9482$. Without scSE, recall increased to $0.9522$, but precision decreased to $0.9238$ and specificity decreased to $0.8953$. This indicates a greater tendency to include surrounding tissue when feature recalibration is removed. 
The ablation variants remained competitive because most of the proposed architecture was still present. Their strong performance therefore shows that DualMiT-Net is not dependent on one isolated component. Instead, the results indicate that the model performance comes from the combined use of local detail, global context, feature sharing, feature recalibration, and decoder fusion.

\subsection{Clinical Implications}
\label{subsec:clinical}

Accurate lesion outlines can support Computer-Aided Diagnosis (CAD) systems for mammography. A predicted mask can be used to estimate lesion size, describe lesion shape, or display a candidate boundary for radiologist review~\cite{burt2018deep,mckinney2020international}. DualMiT-Net uses both the lesion region and the surrounding breast tissue, allowing the segmentation decision to use information from more than one field of view.  
The current model assumes that the approximate mass location is already known. Its most direct application would therefore be after a suspicious lesion has been identified by a radiologist or an automated detection model. DualMiT-Net could then provide a detailed segmentation of that region for subsequent analysis.

\subsection{Limitations}
\label{subsec:limitations}

Several limitations should be considered when interpreting these results. 
First, the current evaluation is lesion-centered. The local ROI is obtained using the expert annotation, and the annotation is also used to normalize lesion orientation and scale. The model therefore performs segmentation after the approximate lesion region has been identified. The reported results should not be interpreted as end-to-end full-mammogram detection and segmentation performance. 
Second, all experiments were performed on CBIS-DDSM, which contains digitized screen-film mammograms. Evaluation on independent datasets and full-field digital mammography is needed to determine how well the model generalizes across imaging systems and patient populations. 
Third, the reported models were trained without random data augmentation. The current experiments were designed to compare the architecture and input representation under the same training conditions. The effect of augmentation can therefore be examined separately in future work. 
Finally, some segmentation errors remain in low-contrast lesions and cases with unclear boundaries, as shown in Fig.~\ref{fig:failure_cases}. Future work will focus on automatic lesion localization, evaluation on independent mammography datasets, and further improvement of boundary segmentation in difficult cases.

\section{Conclusion}
\label{sec:conclusion}

This paper presented DualMiT-Net, a dual-branch encoder-decoder model for breast mass segmentation in lesion-centered mammographic ROIs. The model combines a local lesion view processed by a MiT-B5 encoder with a wider contextual view processed by an EfficientNet-B5 encoder. Features from the two branches are shared at the deepest levels and progressively fused in a single decoder. The selected input representation combines percentile-windowed mammograms with Gabor texture information.
On the mass subset of CBIS-DDSM, DualMiT-Net achieved a Dice coefficient of $0.9375 \pm 0.0011$ and an IoU of $0.8834 \pm 0.0020$ across three training runs. The proposed model outperformed six standard segmentation baselines evaluated using the same dataset and training settings and achieved the highest Dice and IoU values among the CBIS-DDSM methods included in our literature comparison. The results support the use of complementary local and global views for preserving lesion boundary detail while retaining information from the surrounding breast tissue.

\end{document}